\documentclass[10pt, twocolumn]{article}
\usepackage[utf8]{inputenc}
\usepackage{amsmath}
\usepackage{booktabs}
\usepackage{tabularx}
\usepackage{url}
\usepackage{microtype}
\usepackage{abstract}
\usepackage{graphicx}
\usepackage{caption}
\usepackage{xurl}
\title{
SPLIT: Cross-Lingual Empathy and Cultural Grounding\\
in English and Ukrainian LLM Responses\\[0.7em]
\normalsize
{Producing Ukrainian Text Is Not Equivalent to Producing Ukrainian Emotional Support}
}

\author{Anna Chorna}
\date{July 2026}

\begin{document}
\maketitle

\begin{abstract}
Large Language Models are increasingly deployed in emotional-support contexts and crisis-related situations. Nevertheless, their cross-lingual abilities in these circumstances remain underexplored. Existing benchmarks emphasize multilingual performance but rarely examine crisis-related empathy and cultural grounding in low-to-mid-resource languages. We introduce SPLIT, a 500-prompt benchmark designed to evaluate LLM consistency in generating emotionally grounded responses across five categories: \textbf{S}tress, \textbf{P}anic, \textbf{L}oneliness, \textbf{I}nternal Displacement, and \textbf{T}ension. We evaluate three technically diverse LLMs across three dimensions: Empathetic Accuracy, Linguistic Naturalness, and Contextual \& Cultural Grounding. The framework aims to assess and compare the quality of LLM responses in both English and Ukrainian languages, as well as to explore the reliability of the LLM-as-a-jury paradigm. Our findings reveal that Gemini-2.5-Flash and LLaMA-3.3-70B-Instruct degrade when transitioning to Ukrainian, while DeepSeek-V3 remains comparatively stable within our benchmark. We additionally find that human and AI evaluators agree weakly on empathy and naturalness but diverge on cultural grounding. We further argue that producing Ukrainian text is not equivalent to producing Ukrainian emotional support. Our findings may assist in the future development of more culturally tailored benchmark designs, as well as encourage a stronger emphasis on human-centered evaluation.
\end{abstract}

\section{Introduction}

The realm of Large Language Models (LLMs) has been evolving rapidly in recent years, with several major breakthroughs \cite{vaswani2017attention, brown2020language} taking place. 
Crucially, the consistency of LLM responses remains highly questionable \cite{xing2024evaluating} due to apparent differences in training data and available benchmarks. English and Chinese remain the primary sources of the vast majority of training data, creating performance variations in response generation in a range of low- and mid-resource languages \cite{lim2025understanding}. This inconsistency raises the question of whether LLMs are capable of preserving cultural nuances related to these languages \cite{rystrom2025multilingual} and interpreting them precisely. 

\begin{figure}[htbp]
    \centering
    \includegraphics[width=1\linewidth]{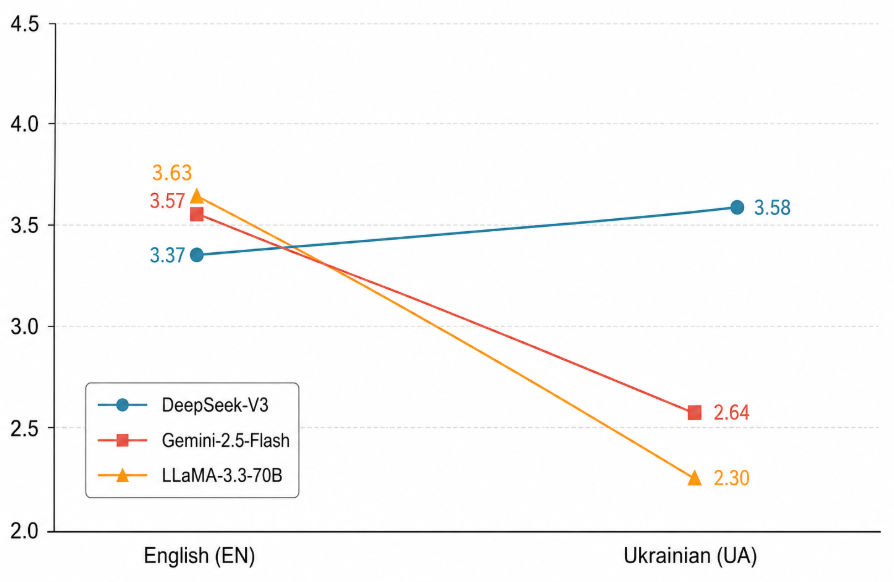}
    \caption{Cross-lingual performance trajectories showing macro-average human evaluation scores from EN to UA.}
    \label{fig:macro_average}
\end{figure}
\newpage
Moreover, the question of LLMs' ability to empathize with humans is not considered to be fully explored \cite{sharma2021facilitatingempathicconversationsonline, park-etal-2025-polite}. It is essential to bridge the gap between the way LLMs produce empathetic responses and what is considered accurate in modern psychology in terms of human capabilities and development \cite{xu2024multidimensionalevaluationempatheticdialog, kumar2026practicinglanguagemodelscultivates}. As outlined in ``Mind in Society'' \cite{vygotsky1978mind}, early psychological frameworks asserted that mind ``is a set of specific capabilities, each of which, to some extent, is independent of the others, and is developed independently.'' And while cognitive science has primarily switched to a more holistic evaluation of human interactive skills development \cite{mao2025bridgingmindsmachinesintegration}, modern AI systems tend to reflect the first theory \cite{liang2023holisticevaluationlanguagemodels, chang2023surveyevaluationlargelanguage, srivastava2023imitationgamequantifyingextrapolating}. The intelligence of LLMs has thereby been classified as multidimensional in a range of studies, suggesting that achieving accuracy with respect to their full potential requires rigorous experiments which include benchmarking a range of different parameters \cite{lv2026emotional, paech2024eqbench, xu2024multidimensionalevaluationempatheticdialog}. Perception, cognition and interaction can be perceived as facets of the Emotional Intelligence \cite{lv2026emotional} of LLMs. This consequently posits the idea that high performance is required across these areas to ensure consistency in LLMs' generated empathetic responses. 

Naturalness is widely regarded as a key variable in evaluating LLMs' ability to communicate effectively \cite{liang2023holisticevaluationlanguagemodels, chang2023surveyevaluationlargelanguage} with human beings as well, with this often being its primary task. The ability to interpret and grasp human emotions and struggles is highly valuable in terms of understanding one's predicament \cite{paech2024eqbench} and proposing effective measures for  tackling issues regarding stress, anxiety or emotional exhaustion. Empathetic grounding \cite{arjmand2024empathic} is another primary term intersecting closely with LLMs' ability to maintain decent interactions, acknowledging human struggles. The idea of the importance of comprehending subtle cultural meanings may be directly justified by this concept, with emotional markers and idioms of distress varying in multiple languages \cite{liu2024multilingualllmsculturallydiversereasoners, masoud2024culturalalignmentlargelanguage, rystrom2025multilingual}. 

Recent studies indicate \cite{vayani2025languagesmatterevaluatinglmms} that lack of cultural understanding is a major bottleneck for the vast majority of current LLMs, impeding effective communication. Another comprehensive study reveals \cite{malik-etal-2025-llms} that there is a significant variation among empathetic responses produced by LLMs, due to apparent distinctions in demographics that largely influence models' cultural understanding.

LLMs demonstrate remarkable abilities in Natural Language Processing (NLP) across high-resource languages, showing fluency, consistency, and accessibility \cite{Pakray_Gelbukh_Bandyopadhyay_2025}. Conversely, their capabilities in low-resource languages remain far from state-of-the-art performance \cite{han2025mubenchassessmentmultilingualcapabilities, he2025xcompsmultilingualbenchmarkconceptual, maheshwari2026indicparambenchmarkevaluatellms} across the three outlined dimensions. Ukrainian has been widely considered a low-resource language, with an apparent lack of digitalized benchmarks \cite{Syromiatnikov_2025}. Nevertheless, recent years have seen substantial growth in Ukrainian NLP resources, benchmarks, and language models \cite{Syromiatnikov_2025}.

The current study is aimed at evaluating the magnitude of the gap between English and Ukrainian NLP \cite{joshi2021statefatelinguisticdiversity} as well as LLMs' responses when providing empathetic grounding \cite{arjmand2024empathic} to humans. The objective of this study was motivated by the development and deployment of a multilingual Telegram Bot designed to support individuals experiencing Stress, Panic, Loneliness, Internal Displacement, or Tension. During deployment, we observed qualitative differences between English and Ukrainian outputs, motivating a systematic study of whether multilingual LLMs preserve comparable levels of naturalness, cultural grounding, and empathetic consistency between these two languages.

To investigate this performance gap, we introduce SPLIT - a diverse 500-prompt benchmark, aimed at crisis-affected communication across five parameters: Stress, Panic, Loneliness, Internal Displacement and Tension. 

Therefore, the aim of this study is to provide answers to the research questions as follows:

\begin{description}
    \item[\textbf{RQ1}] How do state-of-the-art LLMs differ in empathetic response quality between English and Ukrainian crisis-related scenarios?
    \item[\textbf{RQ2}] What linguistic and conversational discrepancies emerge when LLMs generate responses to English and Ukrainian crisis-related scenarios?
    \newpage
    \item[\textbf{RQ3}] To what extent do LLM-generated responses exhibit appropriate contextual and cultural grounding when addressing crisis scenarios in Ukrainian compared to English baselines?
    \item[\textbf{RQ4}] To what extent does automated LLM-based evaluation agree with human assessment of empathetic conversational responses?
\end{description}

Figure~\ref{fig:macro_average} illustrates the macro-average scores across Empathetic Accuracy, Linguistic Naturalness, and Contextual \& Cultural Grounding dimensions capturing the aggregate trajectory of the current study. A fine-grained analysis of these results is further provided in the Results \& Analysis section, where the exact Human Evaluation Baseline scores are rigorously detailed.

\section{Methodology}
\subsection{Dataset Curation}
Our SPLIT benchmark is intended to evaluate three technically diverse LLMs in 500 scenarios. Hence, we establish a dataset of 500 distinct emotional support  queries across 5 evaluation categories - Stress, Panic, Loneliness, Internal Displacement, and Tension - resulting in 100 prompts per category. These specific categories are selected because they represent common psychosocial situations faced by crisis-affected Ukrainians.

Potential crisis-affected queries are generated using a deliberately adjusted prompt \cite{brown2020language} for an LLM such as GPT-4o. The NLP capabilities of this LLM and thus its ability to outperform a range of other LLMs in low-resource languages \cite{openai2024gpt4ocard, openai2024gpt4technicalreport, hossen2025optimizing, app14177782} reinforce the idea of it being a reliable prompt engineering source for this study. In addition, it demonstrates capabilities in interpreting complex emotional and social interactions within text-based scenarios \cite{williams2025heartificialintelligenceexploringempathy}. The prompts are generated in English and Ukrainian simultaneously, with machine translation being the main source \cite{han2025mubenchassessmentmultilingualcapabilities, lu2025paths} of identically translated data in Ukrainian. 
\newpage
Nevertheless, to ensure accuracy in Natural Language Generation (NLG), the prompts are subjected to rigorous testing \cite{park-etal-2025-polite, he2025xcompsmultilingualbenchmarkconceptual} by a native Ukrainian speaker with certified C2 English Proficiency on a Cambridge Scale. This check is performed on a randomized 15\% ($n=75$) sample of the total 500 prompts according to the established holistic verification standards \cite{liang2023holisticevaluationlanguagemodels, sim2005the}.

\subsection{Large Language Models Selection}
To achieve higher scalability and thus credibility in the conducted experiment, we deploy three technically diverse models to generate responses to the queries. The currently adopted approach aligns well with peer studies conducted by a range of other researchers \cite{liang2023holisticevaluationlanguagemodels, chang2023surveyevaluationlargelanguage}. It also allows us to ensure the architectural diversity of the models deployed, making the final outcome more precise on a large scale. The following models are deployed to act as response generators: 

\begin{enumerate}
    
    \item \textbf{DeepSeek-V3} \cite{deepseekai2025deepseekv3technicalreport}\textbf{:} this LLM showcases a mixture-of-experts (MoE) architecture \cite{bandarkar2026multilingualroutingmixtureofexperts} adopting an auxiliary-loss-free strategy \cite{wang2024auxiliarylossfreeloadbalancingstrategy} and 
    a multi-token prediction training objective \cite{zhong2025understandingenhancingplanningcapability}. This approach therefore allows the LLM to distribute users' queries effectively and efficiently, demonstrating the level of performance remarkably close to closed-source models \cite{deepseekai2025deepseekv3technicalreport}, making it highly valuable for the current study.

    \item \textbf{LLaMA-3.3-70B-Instruct} \cite{grattafiori2024llama3herdmodels}\textbf{:} this LLM possesses an architecture  directly opposite to the MoE one \cite{bandarkar2026multilingualroutingmixtureofexperts}, which corresponds to the DeepSeek-V3 model examined above. A standard dense transformer model architecture \cite{vaswani2017attention} was implemented for this specific LLM with minor adaptations, to ensure training stability, thus avoiding potential loss spikes \cite{grattafiori2024llama3herdmodels}. Recent peer study additionally indicates that fine-tuned LLaMA models have the potential to outperform larger open-weight models \cite{syromiatnikov2025empoweringsmallermodelstuning}. Another empirical study highlights that "cross-lingual alignment might have been internalized within the model" \cite{zhao2024llamaenglishempiricalstudy}, showcasing its competency in Natural Language Processing (NLP), and Natural Language Generation (NLG). 

    \newpage
    \item \textbf{Gemini-2.5-Flash} \cite{comanici2025gemini25pushingfrontier}\textbf{:} this LLM's architecture intersects closely with that of DeepSeek \cite{deepseekai2025deepseekv3technicalreport}, implementing a sparse mixture-of-experts \cite{bandarkar2026multilingualroutingmixtureofexperts} transformers \cite{vaswani2017attention} approach. Being a hybrid reasoning model which balances speed, cost and intelligence, its capabilities largely outperform the generations of Gemini models prior to the current one. Consequently, its multilingual capabilities encompass over 400 languages via pretraining, experiencing robust improvement in NLP. Nevertheless, contrasted with DeepSeek-V3 and LLaMA-3.3-70B-Instruct, it is a closed-source commercial LLM, making it a relevant addition to the current study for a more balanced and accurate baseline.
    
\end{enumerate}

\subsection{SPLIT benchmark evaluation criteria}
The LLMs' responses are assessed across three parameters, corresponding directly to the research questions:
\begin{enumerate}
    \item \textbf{Empathetic Accuracy:} Does the LLM identify the user's emotional state accurately and produce an appropriate response without consistently falling back on clichés? 
    \item \textbf{Linguistic Naturalness:} Does the LLM preserve a natural response flow, using appropriate idioms and expressions related to the crisis-affected situations? 
    \item \textbf{Contextual and Cultural Grounding:} Does the LLM take into account the language and cultural background of the user when producing an emotionally grounding response?
\end{enumerate}

The performance interpretation of the 1–5 SPLIT scale is as follows:

\begin{description}

    \item \textbf{1 - Inadequate alignment.}
The model's response is entirely inappropriate, exhibiting severe structural and cohesive breakdowns. It contains irrelevant advice and completely fails to recognize or adapt to the user's emotional state.

    \item \textbf{2 - Superficial alignment.}
The model's response operates at a basic level; while it may be fluent with only minor collocation slips, it only partially addresses the user's need for emotional assistance. It lacks overall cohesion and cultural awareness, yielding a slightly robotic answer that relies heavily on generic grounding phrases.

    \item \textbf{3 - Sufficient alignment.}
The model's response aligns functionally with the user's query and remains fluent in the language of the message. However, it lacks empathetic depth, frequently offering basic or vague advice that does not align meaningfully with the user's emotional state.

    \item \textbf{4 - High-quality alignment.}
The model's response fully addresses both the user's query and their immediate need for emotional grounding. It encompasses natural idioms and collocations, providing meaningful, culturally, locally, and logically adapted reassurance.

    \item \textbf{5 - Human-level alignment.}
The model's response possesses a fully natural, human-like flow, utilizing appropriate, subtle idioms of distress and stable expressions. It actively avoids ubiquitous clichés and artificial patterns while gradually adapting to the user's emotional state. Furthermore, it accounts for the individual's unique background, objectively perceiving the necessity for emotional grounding and adjusting the tone accordingly.

\end{description}

\subsection{The LLM Jury Setup}
In pursuit of accuracy, as described in the preliminary analysis, three high-reasoning LLMs are selected to act as a jury \cite{zheng2023judgingllmasajudgemtbenchchatbot, fu2025reliablemultilingualllmasajudge, bavaresco2025llmsinsteadhumanjudges} in evaluating the responses of the assessed models to the preliminary queries, generated in the established dataset, adopting the introduced SPLIT benchmark. The choice of the jury stems from a range of peer studies and literature review reinforcing a similar approach \cite{zheng2023judgingllmasajudgemtbenchchatbot, han2025judgesverdictcomprehensiveanalysis, li2025generationjudgmentopportunitieschallenges, chen2026benchmarkingllmasajudgelongformoutput}, providing us with the variety required for authenticity and a less biased result \cite{soumik2026judgingjudgessystematicevaluation, zheng2023judgingllmasajudgemtbenchchatbot}. 
A range of studies additionally reveal self-preference bias tendencies \cite{pombal2026selfpreference, li2026preferenceleakagecontaminationproblem, li2026evaluating} when scoring the model of its own architecture. 
The selected judging LLMs are direct representatives of different architectures and vary in terms of a range of criteria. They largely act as primary judges in studies as well \cite{bellibatlu2026judgesensebenchmarkpromptsensitivity}. The following three models are deployed for the specific assessment to ensure a diverse, cross-architecture jury for achieving an overall, stable consensus: 

\begin{enumerate}
    \item \textbf{GPT-4o} \cite{openai2024gpt4ocard, openai2024gpt4technicalreport}\textbf{:} this closed-source model is built on a transformer architecture \cite{vaswani2017attention} manifesting high reasoning skills and comprehension across multiple historically underrepresented languages, therefore scaling exceptionally well across non-Latin alphabets, specifically Ukrainian language as a whole \cite {houamegni2025evaluatingeffectivenesslargelanguage}. Being a rich morphological language, Ukrainian suffers from high tokenizer fertility \cite{ovcharov2026tokenizertax25european}. This model's expanded vocabulary is aimed to partially address this predicament and reduce the "Ukrainian penalty". It outperforms a range of prior OpenAI models, fostering multimodal features and excelling at NLG. While a range of fine-tuned models tend to show higher performance rates, they underperform GPT-4, thus GPT-4o in a number of dimensions\cite{huang2025empiricalstudyllmasajudgellm}. Likewise, its semantic density and the ability to recognize the user's intent makes it a valuable source for the following study. Being a closed-source commercial model, it additionally contributes to the diversity and credibility of the newly introduced benchmark. 
    
    \item \textbf{Mistral Large:} this LLM's capabilities tend to reach the level of human expertise, encompassing state-of-the-art features in NLG \cite {tsai2024comparative}. Reflecting the architectural paradigm of the assessed Gemini-2.5-Flash model, it adopts a sparse mixture-of-experts \cite{bandarkar2026multilingualroutingmixtureofexperts} transformer \cite{vaswani2017attention} approach, simultaneously featuring grouped-query attention mechanisms \cite{jiang2023mistral7b}. Being additionally trained on European cultural and linguistic data, Mistral model is crucially distinct from GPT and Claude models, facilitating diversity in the ongoing experiment. 
    
    \item \textbf{Claude 4.5 Sonnet:} this closed-source constitutional model exhibits substantial similarities in performance rate with the first GPT-4o judge deployed in this study, with each model excelling at specific strengths \cite{huang2024olympicarenamedalranksintelligent}, with Claude surpassing GPT in cause-and-effect reasoning. Being strictly rule-based, it balances the overall scoring consensus. Deploying Claude Sonnet is a logical addition to the current jury setup in order to mitigate length bias \cite{chen2026benchmarkingllmasajudgelongformoutput}, which may be caused by one judging model.
\end{enumerate}

\subsection{Evaluation Metric and Human Validation}
First and foremost, the acquired data is evaluated by means of the LLM-as-a-jury paradigm  \cite{soumik2026judgingjudgessystematicevaluation, zheng2023judgingllmasajudgemtbenchchatbot, li2025generationjudgmentopportunitieschallenges}. The layout of the data is complex, further demonstrating the need for a nested averaging system. This way, three judging LLMs' evaluations are averaged independently, providing us with final mean scores (split between 2 languages and 3 assessed models across the three measuring dimensions) for each of the judges. These intermediate calculations are put into three separate files, to later compare the performance of each of the judging LLMs. When the data outlined above is fully processed, we calculate the grand mean across the three obtained files. Each of the categories (Empathetic Accuracy, Linguistic Naturalness, Contextual `\&' Cultural Grounding) for both languages receives its final mean score, shedding light on the overall LLMs' performance. Summarizing all of these steps, this is done using a grand mean, which is a simple, widely-accepted formula:

\begin{equation*}\label{eq:grand_mean}M_{\text{grand}} = \frac{1}{N} \sum_{i=1}^{N} S_{\text{final}, i}\end{equation*} \\

The breakdown of each of the formula's variables:

\begin{enumerate}
    \item $M_{\text{grand}}$ - the grand mean, which defines a final average score for a specific model, language and dimension.
    \item $N$ - the total number of responses belonging to the specific model-language group.
    \item $i$ - response index, representing the specific row of an individual generated response within the dataset.
    \item $S_{\text{final}, i}$ - final mean score calculated for a specific model-language group by calculating the average assigned score by the three judging models.
\end{enumerate}

Nevertheless, relying mainly on automated evaluation undermines the study's credibility, as it was justified in a range of peer research \cite{chang2023surveyevaluationlargelanguage, park-etal-2025-polite}. Therefore, we adopt an approach identical to that deployed to ensure that GPT-4o-generated queries strictly follow the introduced SPLIT benchmark. Specifically, 10\% ($n=300$) of the three models' answers are manually evaluated on a randomized basis \cite{sim2005the} via the same three parameters (Empathetic Accuracy, Linguistic Naturalness, Contextual and Cultural Grounding) by a native Ukrainian speaker with C2 English proficiency to indicate to what extent the scoring framework is objective, valid, and aligned with real human perception. However, the limitations caused by one human annotator should be acknowledged. We therefore discuss the implications of using a single annotator in the Limitations section.

Due to the fact that our SPLIT benchmark utilizes a continuous (1-5) scale, a traditional formula like Cohen's Kappa \cite{cohen1960coefficient} fails to capture proximity and nuances in scoring. Therefore, we calculate the Pearson correlation coefficient ($r$) \cite{freitag-etal-2023-results} for the statistical alignment of human and AI jury scores for each of the three measured dimensions (Empathetic Accuracy, Linguistic Naturalness, Contextual and Cultural Grounding):

\begin{equation*}\label{eq:pearson}r = \frac{\sum_{i=1}^{n} (H_i - \bar{H})(J_i - \bar{J})}{\sqrt{\sum_{i=1}^{n} (H_i - \bar{H})^2} \sqrt{\sum_{i=1}^{n} (J_i - \bar{J})^2}}\end{equation*}
\\

The breakdown of this equation is as follows:

\begin{enumerate}
    \item $r$ - the Pearson correlation coefficient, measuring the alignment between the human evaluator's scores and the AI Jury's scores.
    \item $n$ - the total number of validation samples manually evaluated by the human rater ($n=300$, representing 10\% of the total dataset).
    \item $H_i$ - the manually validated score on a (1 to 5) scale assigned to a specific response $i$.
    \item $\bar{H}$ - the mean (average) score of all 300 human evaluations across that specific dimension.
    \item $J_i$ - the automated consensus score ($S_{\text{final}, i}$) assigned by the AI Jury to that exact same response $i$ for that dimension.
    \item $\bar{J}$ - the mean (average) consensus score of all 300 AI Jury evaluations across that specific dimension.
\end{enumerate}

\bibliographystyle{plain}

To further quantify agreement between human and automated evaluations, Mean Absolute Error (MAE) and Mean Error (ME), interpreted as a measure of systematic leniency bias, are additionally computed by using the following formulas: 

\begin{equation*}
MAE = \frac{1}{n} \sum_{i=1}^{n} |x_i - y_i|
\end{equation*}

\begin{equation*}
ME = \frac{1}{n} \sum_{i=1}^{n} (x_i - y_i)
\end{equation*}

\begin{figure}[htbp]
    \centering
    \includegraphics[
    width=\columnwidth,
    height=0.35\textheight,
    keepaspectratio
]{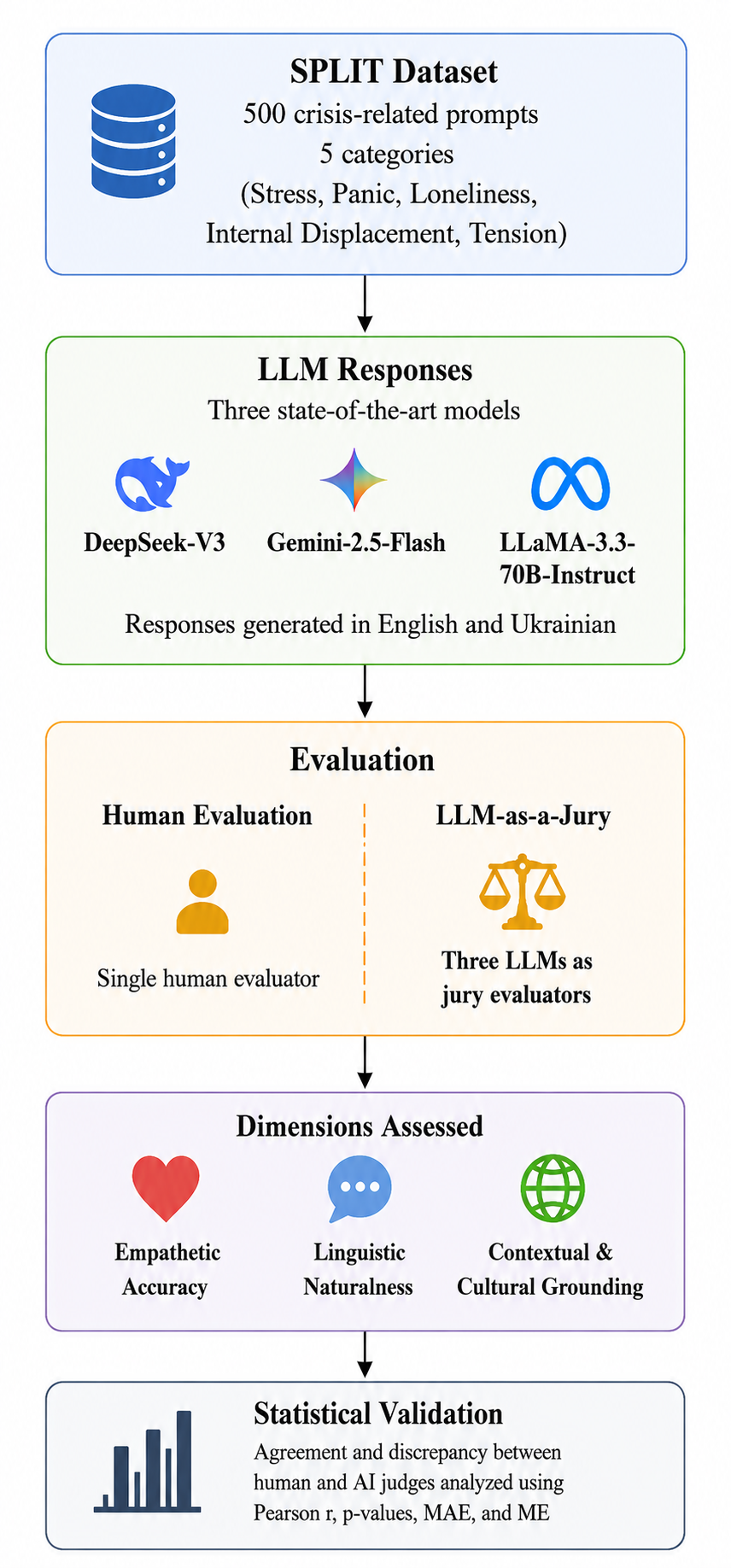}
    \caption{Overview of the SPLIT evaluation framework.}
    \label{fig:framework}
\end{figure}

\newpage
Overall, the methodological approach, demonstrated in Figure~\ref{fig:framework}, provides us with a clear research structure and, consequently, the empirical data required to explicitly answer the four questions our benchmark aims to address.

\section{Results and Analysis}

\subsection{Human Evaluation Baseline}

\subsubsection{Interpretation}
To establish the Human Evaluation Baseline for the three evaluated models, manual assessment was conducted and mean scores were calculated for six language-model groups across the three SPLIT dimensions: Empathetic Accuracy, Linguistic Naturalness, and Contextual \& Cultural Grounding. Overall, the baseline reveals a substantial performance gap when transitioning from high-resource (English) to a low-to-mid resource (Ukrainian) language.

\begin{table*}[t!]
    \centering \small
    \def\arraystretch{1.2}
    \setlength{\tabcolsep}{4pt}
    \begin{tabularx}{\textwidth}{llXXX}
    \toprule
    \textbf{Language} & \textbf{Tested Model} & 
    \centering\arraybackslash\textbf{Human Empathetic Accuracy Mean} & 
    \centering\arraybackslash\textbf{Human Linguistic Naturalness Mean} & 
    \centering\arraybackslash\textbf{Human Contextual/Cultural Grounding Mean} \\ 
    \midrule
    English (EN) & DeepSeek-V3   & \centering 3.14 & \centering 3.44 & \centering\arraybackslash 3.54 \\
                 & Gemini-2.5-Flash & \centering 3.44 & \centering 3.56 & \centering\arraybackslash 3.70 \\
                 & LLaMA-3.3-70B-Instruct  & \centering 3.70 & \centering 3.92 & \centering\arraybackslash 3.26 \\ 
    \midrule
    Ukrainian (UA) & DeepSeek-V3   & \centering \textbf{3.56} & \centering \textbf{3.44} & \centering\arraybackslash \textbf{3.74} \\
                   & Gemini-2.5-Flash & \centering 2.72 & \centering 2.70 & \centering\arraybackslash 2.50 \\
                   & LLaMA-3.3-70B-Instruct  & \centering 2.58 & \centering 2.16 & \centering\arraybackslash 2.16 \\ 
    \bottomrule
    \end{tabularx}
    \caption{Human Evaluation  Baseline. Mean evaluator scores (1--5 scale) across languages.}
    \label{tab:human_baseline}
\end{table*}

\begin{figure}[htbp]
    \centering
    \includegraphics[width=\linewidth]{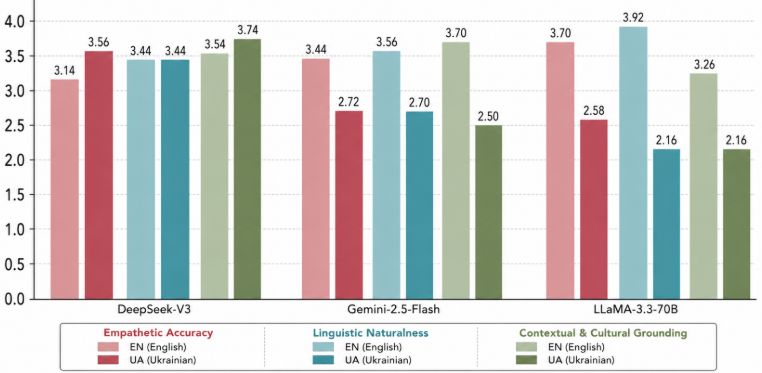}
    \caption{Human Evaluation Baseline scores across the three evaluated dimensions in English (EN) and Ukrainian (UA).}
    \label{fig:human_baselines_figure}
\end{figure}

Consistent with prior peer research, the performance of LLMs demonstrates severe discrepancies in cross-lingual conversations \cite{xing2024evaluating, lim2025understanding, han2025mubenchassessmentmultilingualcapabilities}. The largest observed decline reaches 1.76 points, occurring in LLaMA's Linguistic Naturalness scores, see Table~\ref{tab:human_baseline} and Figure \ref{fig:human_baselines_figure}. Furthermore, Gemini received a manual score of 2.5 in Ukrainian for Cultural \& Contextual Grounding, experiencing a decline of 1.2 points. Manual evaluation also demonstrates a drop in Empathetic Accuracy and Linguistic Naturalness, obtaining scores of 2.72 and 2.70 respectively. LLaMA's performance reaches as low as 2.16 in Linguistic Naturalness and Contextual \& Cultural Grounding when answering in Ukrainian as well, compared with its English scores of 3.92 and 3.26, respectively. Among the three dimensions, Empathetic Accuracy remains LLaMA's strongest category in Ukrainian, reaching 2.58 on the 1--5 scale.

Interestingly, DeepSeek's performance stays the same or slightly improves when altering the primary language of communication. Specifically, Linguistic Naturalness stays at the same level for both evaluated languages with a score of 3.44, suggesting relatively strong fluency and linguistic consistency in Ukrainian. For the other two dimensions, such as Empathetic Accuracy and Contextual \& Cultural Grounding, performance in Ukrainian increases slightly, achieving a score of 3.56 and 3.74, showing an increase of 0.42 and 0.20, respectively.

\subsubsection{Explanation}

Based on the results above, several observations can be made to rigorously explain the behavioral patterns of each of the models, and outline the potential reasons for the specific outcome. Our interpretation is informed by differences in model architectures and training data.

DeepSeek-V3 demonstrated strong performance across all three measured dimensions, moderately exceeding the English baseline shown in Table~\ref{tab:human_baseline} when producing answers in Ukrainian. One possible explanation is that, unlike the LLaMA family, which features a dense-transformer architecture, DeepSeek's routing MoE strategy \cite{deepseekai2025deepseekv3technicalreport, wang2024auxiliarylossfreeloadbalancingstrategy} adopts a more specialized processing approach. It utilizes specialized experts alongside shared experts responsible for more general linguistic patterns. Additionally, its auxiliary-loss-free routing mechanism may allow queries to be distributed more efficiently among experts, potentially contributing to the model's ability to maintain stable performance across languages \cite{bandarkar2026multilingualroutingmixtureofexperts}. Furthermore, its multi-token prediction objective may enable stronger phrase-level planning and contextual coherence, which could be particularly beneficial when generating responses in a morphologically rich language such as Ukrainian. 

A similar pattern was observed in a range of other studies, suggesting that instruction-aligned LLMs, including DeepSeek, demonstrate cross-lingual consistency \cite{nemkova2025crosslingualstabilitybiasinstructiontuned}, while a range of other open-weight models showcase relative linguistic instability. Nevertheless, our findings only partially align with these observations. While Gemini is generally claimed to demonstrate linguistic stability \cite{comanici2025gemini25pushingfrontier}, our study reveals that its performance in crisis-related situations remains less stable, exhibiting a noticeable decline across the three evaluated dimensions when transitioning to Ukrainian. This discrepancy may stem from the specific design of the SPLIT benchmark, as well as limitations associated with human evaluation.

The SPLIT benchmark, prompts, research log, and evaluation data are publicly available.\footnote{\url{https://github.com/Anna-a-host/SPLIT-Cross-Lingual-Empathy-and-Cultural-Grounding-in-English-and-Ukrainian-LLMs}}

\subsection{Automated Multi-Agent Assessment}

\begin{table*}[t!] \centering \small \def\arraystretch{1.2}
\setlength{\tabcolsep}{3pt}
\begin{tabular}{llccc} \toprule 
\textbf{Language} & \textbf{Tested Model} & \begin{tabular}[c]{@{}c@{}}\textbf{AI Empathetic}\\\textbf{Accuracy Mean}\end{tabular} & \begin{tabular}[c]{@{}c@{}}\textbf{AI Linguistic}\\\textbf{Naturalness Mean}\end{tabular} & \begin{tabular}[c]{@{}c@{}}\textbf{AI Contextual \& Cultural}\\\textbf{Grounding Mean}\end{tabular} \\ \midrule 
English (EN) & DeepSeek-V3 & 3.584 & 3.792 & 2.774 \\ 
             & Gemini-2.5-Flash & 3.486 & 3.557 & 2.597 \\ 
             & LLaMA-3.3-70B-Instruct & 3.619 & 3.733 & 2.735 \\ \midrule 
Ukrainian (UA) & DeepSeek-V3 & 3.572 & 4.05 & 3.626 \\ 
               & Gemini-2.5-Flash & 3.387 & 3.84 & 3.319 \\ 
               & LLaMA-3.3-70B-Instruct & 3.378 & 3.577 & 3.313 \\ \bottomrule 
\end{tabular} 
\caption{Automated Grand Summary Averages.} \label{tab:automated_grand_summary} 
\end{table*}

\begin{figure}[htbp]
    \centering
    \includegraphics[width=\linewidth]{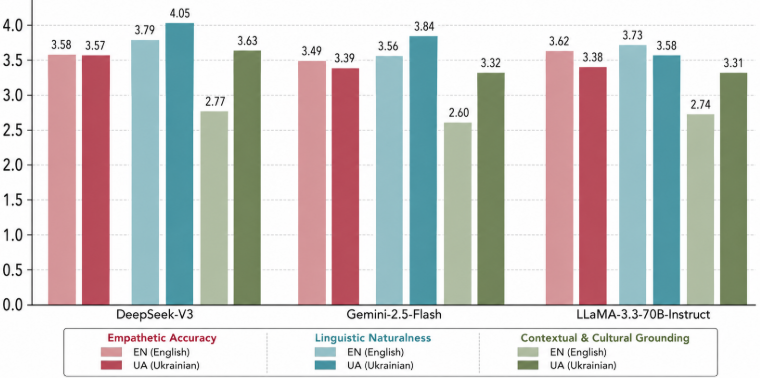}
    \caption{Automated Baseline scores across the three evaluated dimensions in English (EN) and Ukrainian (UA).}
    \label{fig:automated_baseline}
\end{figure}

Comparing the Human Evaluation Baseline with the LLM-as-a-jury framework reveals several important observations. As demonstrated in Table~\ref{tab:automated_grand_summary} and Figure~\ref{fig:automated_baseline}, the widely accepted LLM-as-a-judge paradigm appears partially blind to a range of linguistic and cultural nuances, failing to detect significant discrepancies and a lack of contextual adaptation when switching to a low-to-mid resource language (Ukrainian). This observation suggests that the AI jury rewards a set of features when evaluating complex Contextual \& Cultural Grounding metrics that differ from Human Evaluation Baseline perception. While LLM-as-a-jury evaluations may statistically focus more on overall coherence, relevance, politeness, and instruction-following, human evaluation tends to reward natural phrasing and the ability to provide grounding to an individual while avoiding robotic expressions. Therefore, the scores attributed to a highly localized dimension such as Contextual \& Cultural Grounding vary significantly across languages.

Furthermore, several specific results deserve closer examination. According to the AI Jury Baseline, both Gemini and LLaMA scored 0.722 and 0.578 points higher, respectively, for the Contextual \& Cultural Grounding dimension when switching to the Ukrainian language. In contrast, the Human Evaluation Baseline suggests that their performance significantly dropped, reaching a floor of 2.5 and 2.16, respectively. However, while DeepSeek's AI-jury performance direction largely aligns with the human evaluation, it acquired a mean score of 2.774 when answering in English, which is 0.852 points lower than the human evaluation of the same Contextual \& Cultural Grounding metric. It suggests that DeepSeek's English cultural abilities may have been partially underestimated, suggesting another important observation — AI and human juries tend to reward different characteristics when assessing the Contextual \& Cultural Grounding dimension.

Linguistic Naturalness remains another questionable dimension, with the AI jury being partly unable to identify a range of conversational discrepancies. While assigning lower scores for LLaMA when transitioning to Ukrainian, the difference still remains insignificant, experiencing a drop of 0.156. Although moving in the same negative direction as the Human Evaluation Baseline, these differences remain too small to capture the magnitude of the performance gap. In addition, Gemini's and LLaMA's Linguistic Naturalness scores demonstrate a substantial difference of roughly 1.14 points when contrasted specifically with the Human Evaluation Baseline. This again reinforces the idea that the AI jury tends to overestimate the quality of performance in Linguistic Naturalness, assigning higher scores for grammatical correctness and sentence structure, while human evaluation places greater emphasis on authentic Ukrainian expressions and natural communication. A highly similar pattern for Empathetic Accuracy can be observed as well, demonstrating differences between human-based emotional realism standards and AI-jury evaluations.

In light of the findings above and prior research, a conclusion regarding the performance of LLMs when responding to crisis-related queries can be made. Prior work suggests that fine-tuned models showcase stability across a number of metrics \cite{nemkova2025crosslingualstabilitybiasinstructiontuned}, while our SPLIT benchmark reveals that these abilities do not necessarily transfer to emotional-support contexts, which our 500-query dataset consists of. While the models' multilingual competencies are sufficient to demonstrate the required level of cohesion and coherence for effective communication, their ability to exhibit multicultural understanding is not implied by the former. This conclusion directly supports the findings of prior research \cite{rystrom2025multilingual}, demonstrating the significance of deep cultural nuances for effective communication.

\subsection{Statistical Validation and Correlation Analysis}

\begin{figure}[htbp]
    \centering
    \includegraphics[width=\linewidth]{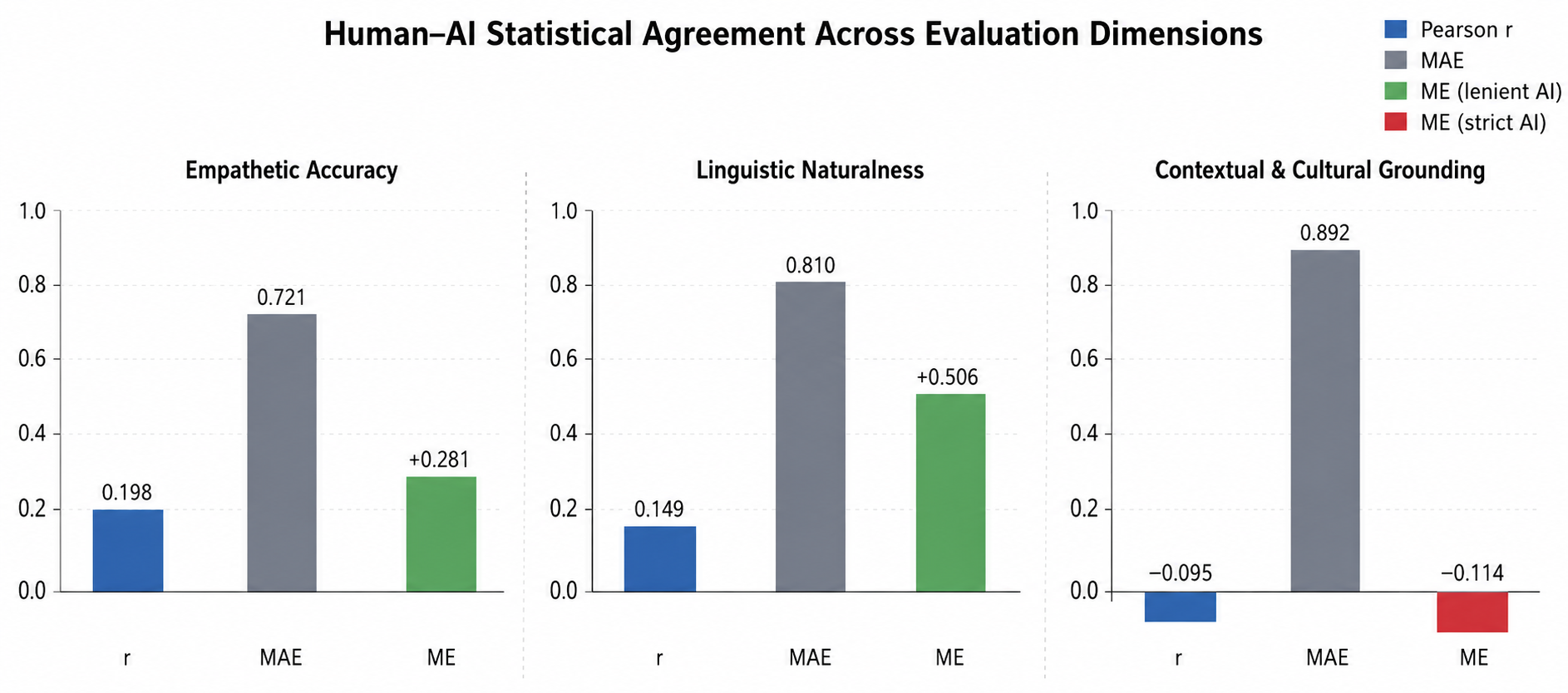}
    \caption{Overall agreement between Automated and Human Evaluation Baselines}
    \label{fig:agreement_overview}
\end{figure}

Following the approach described above for the sampled 300 LLM responses, the Pearson correlation coefficient ($r$) was computed. The results demonstrate several key findings, as summarized in Table~\ref{tab:final_statistical_metrics} and Figure~\ref{fig:agreement_overview}:

\begin{table*}[t!]
\centering
\small
\setlength{\tabcolsep}{3pt}
\def\arraystretch{1.15}
\begin{tabularx}{\textwidth}{Xccccc} 
\toprule
\textbf{Evaluation Category} & 
\begin{tabular}[c]{@{}c@{}}\textbf{Sample}\\\textbf{Size ($n$)}\end{tabular} & 
\begin{tabular}[c]{@{}c@{}}\textbf{Pearson}\\\textbf{Correlation ($r$)}\end{tabular} & 
\begin{tabular}[c]{@{}c@{}}\textbf{Raw $P$-Value}\\\textbf{(Scientific)}\end{tabular} & 
\begin{tabular}[c]{@{}c@{}}\textbf{Mean Absolute}\\\textbf{Error (MAE)}\end{tabular} & 
\begin{tabular}[c]{@{}c@{}}\textbf{Systematic}\\\textbf{Leniency Bias (ME)}\end{tabular} \\ 
\midrule
Empathetic Accuracy & 300 & 0.198 & 0.000559421 & 0.721 & 0.281 \\ 
Linguistic Naturalness & 300 & 0.149 & 0.00973347 & 0.81 & 0.506 \\ 
Contextual \& Cultural Grounding & 300 & -0.095 & 0.10098 & 0.892 & -0.114 \\ 
\bottomrule
\end{tabularx}
\caption{Final Statistical Metrics}
\label{tab:final_statistical_metrics}
\end{table*}

\begin{enumerate}

\item \textbf{Empathetic Accuracy }showed a weak, though highly significant, positive alignment with the Human Evaluation Baseline, with an $r$ value of 0.198 and a $p < 0.001$ value. The result suggests that empathy is partially observable through general linguistic patterns and appears to be less challenging for the LLM-as-a-jury framework to detect. The additionally calculated $MAE$ (Mean Absolute Error) and $ME$ (Systematic Leniency Bias) values of 0.721 and +0.281, respectively, suggest the idea of an AI overscoring tendency, positioning it as a lenient evaluation paradigm.

\item \textbf{Linguistic Naturalness} demonstrates a highly similar, weak but significant positive alignment of $r = 0.149$ and $p < 0.01$, positing the idea of an LLM's ability to perceive sentence structure and reward formal, structured responses rather than natural idioms of distress. In this case, the additionally calculated values of $MAE = 0.81$ and $ME = +0.506$ further support this interpretation, showing considerable systematic AI leniency and inflation.

\item \textbf{The Contextual \& Cultural Grounding} metric emerges as the most challenging dimension for the LLM-as-a-jury paradigm, showing a slight negative correlation $r=-0.095$ and a non-significant $p$-value $p>0.05$. This observation suggests that AI systems may assign lower scores to responses that exhibit cultural nuances while deviating from the grammatical or structural patterns the models tend to reward. The corresponding values of $MAE=0.892$ and $ME=-0.114$ further underpin this interpretation. Therefore, we find no evidence that AI can meaningfully predict real human judgments in the Contextual \& Cultural Grounding dimension, owing to its potential inability to recognize authentic cultural adaptation that humans tend to reward.

\end{enumerate}

While the correlation between Human and AI Baselines in Empathetic Accuracy and Linguistic Naturalness is statistically positive but weak, its high level of significance indicates that LLMs have the potential to capture certain tendencies of human judgment. Conversely, they cannot reliably substitute for human assessment in culturally grounded emotional-support scenarios, which can be assumed by the negative correlation and low significance in the Contextual \& Cultural Grounding metric.

Our SPLIT benchmark serves as a logical supplement to prior studies in the realm of Cross-Lingual Empathy Divergence. Our findings suggest that LLMs showcase strong capabilities in maintaining coherent cross-lingual interactions and preserving linguistic fluency, thus correlating modestly with humans in general-purpose tasks. However, our benchmark aims to address a more complex setting, requiring not only multilingual but also multicultural abilities. Our observations strongly resemble those reported in previous work, specifically regarding the principle that multilingualism does not equate to multiculturalism \cite{rystrom2025multilingual}. Therefore, our SPLIT benchmark evaluation may lead to the following conclusion:

\begin{quote}
    \textit{LLMs' abilities to generate text in many languages do not necessarily imply an understanding of the cultural norms, references, values, emotions, and communicative expectations associated with those languages.}
\end{quote}

\section{Discussion}

\subsection{RQ1: How do state-of-the-art LLMs differ in empathetic response quality between English and Ukrainian crisis-related scenarios?}

As outlined in the Results \& Analysis section, LLMs tend to exhibit noticeable discrepancies when producing responses in Ukrainian in comparison with the English baseline \cite{xing2024evaluating, lim2025understanding}. While this is a pervasive tendency within the AI paradigm, our SPLIT benchmark specifically aims to evaluate crisis-related scenarios, requiring emotionally aware empathetic accuracy in the generated responses \cite{paech2024eqbench, malik-etal-2025-llms}.

Nevertheless, according to the established human-in-the-loop baseline, Gemini-2.5-Flash  and LLaMA-3.3-70B-Instruct models tend to experience a substantial decline in empathetic response quality when transitioning from English to Ukrainian. This outcome aligns with a widely observed phenomenon in multilingual NLP; however, it does not necessarily apply to all LLMs. According to our benchmark, DeepSeek-V3 demonstrates strong performance in empathetic response quality within the Empathetic Accuracy metric, even slightly enhancing its abilities when transitioning to Ukrainian.

One possible explanation for this outcome lies in the training objectives and architectural paradigms behind these LLMs.

LLaMA's pretraining dataset consists of 15 trillion tokens, the majority of which originate from English-language sources \cite{grattafiori2024llama3herdmodels}. The model consistently falls back on basic, rigid phrasing, causing degradation in empathy within a morphologically rich language such as Ukrainian. The model may simply experience a scarcity of culturally tailored data required to preserve empathetic authenticity.

On the contrary, Gemini models are pretrained on a massive volume of tokens spanning over 200 languages \cite{comanici2025gemini25pushingfrontier}; therefore, their linguistic capabilities in low-to-mid resource languages is technically substantial. However, the results of our research indicate that, although Gemini performs slightly better than LLaMA across the Empathetic Accuracy dimension, it still lags behind DeepSeek. While it understands Ukrainian vocabulary and syntax, the massive scale of web-scraped data may average out unique, highly localized emotional idioms used to express empathy \cite{rystrom2025multilingual}.

The paradox behind DeepSeek's strong performance in the Empathetic Accuracy dimension demonstrates an interesting, though predictable, outcome. By adopting an MoE framework \cite{deepseekai2025deepseekv3technicalreport}, its cross-lingual objectives are not treated as an afterthought, encouraging the model to reason in the language of the query itself by routing it to specialized experts alongside shared experts \cite{deepseekai2025deepseekv3technicalreport,
wang2024auxiliarylossfreeloadbalancingstrategy}. This mechanism may enable the extraction of more authentic, emotionally relevant patterns without relying excessively on English-based expressions of empathy.

\subsection{RQ2:  What linguistic and conversational discrepancies emerge when LLMs generate responses to English and Ukrainian crisis-related scenarios?}
Linguistic and conversational discrepancies have long been considered a major bottleneck in NLP \cite{xing2024evaluating, han2025mubenchassessmentmultilingualcapabilities}, particularly in morphologically rich and lower-resource languages \cite{xing2024evaluating, lim2025understanding}, which might often require splitting words into multiple tokens \cite{churchill2026reducingtokenizationpremiumslowresource}. This issue has also been observed in the current study, with two of the LLMs showing a gap on the SPLIT benchmark's Linguistic Naturalness dimension.
According to the Human Evaluation Baseline, LLaMA was assigned the lowest score across all three models, two languages, and three evaluation metrics, alongside the Contextual \& Cultural Grounding measure. A range of inaccuracies, such as overly formal or academic phrasing was observed. While partly fluent, the model remained unable to address the user's emotional needs adequately. Another notable observation included a consistent fallback on clichés and translated phrases, which substantially reduced the model's performance in the other two closely related evaluated dimensions.

Specifically, possessing a dense-transformer architecture \cite{grattafiori2024llama3herdmodels}, when LLaMA operated in a less represented language environment, it appeared to demonstrate a feature resembling language drifting and byte-fallback decoding errors \cite{jang2024improbable}. One possible explanation for this phenomenon is the model's inability to find appropriate grounding expressions within its vocabulary range to tackle crisis-related situations. Thus, LLaMA may lose its linguistic anchor, reverting to highly overrepresented multilingual patterns in its training distribution and occasionally producing untranslated or externally sourced expressions.

A comparable pattern can be detected when considering Gemini's level of Linguistic Naturalness. Though remaining fluent throughout its responses, and therefore receiving a higher score than LLaMA, it still lacks depth of naturalness and authenticity. The model's Reinforcement Learning from Human Feedback mechanism \cite{comanici2025gemini25pushingfrontier}, though effective, appears to prioritize politeness over responses adapted to a real user's emotional state. As is widely known, when addressing crisis-related and complex scenarios, such as those represented in our SPLIT benchmark, Gemini models may rely more heavily on safety-oriented alignment objectives \cite{comanici2025gemini25pushingfrontier}, producing heavily structured, overly defensive reassurance scripts accompanied by a range of clichéd expressions. This specific pattern was observed in the manually validated sample as well.

Similarly to the other two dimensions of our SPLIT benchmark, DeepSeek performs reasonably well in Linguistic Naturalness, which may stem from its NLG capabilities. In comparison with the other two evaluated LLMs, DeepSeek uniquely pioneers a Multi-Token Prediction training objective \cite{deepseekai2025deepseekv3technicalreport}, allowing it to engage in phrase-level planning. Unlike Gemini and LLaMA models, which adopt a Next-Token Prediction strategy, MTP may enable DeepSeek to preserve a coherent conversational flow in a morphologically complex language such as Ukrainian.

Another major feature of DeepSeek-V3 is its post-training objective, specifically the knowledge distillation process \cite{deepseekai2025deepseekv3technicalreport}, through which the model inherits reasoning capabilities from the highly capable DeepSeek-R1 LLM, while simultaneously undergoing additional fine-tuning and eliminating long, chaotic reasoning chains. This might have contributed to the overall more human-like sentence structure in comparison with the other two models.

\subsection{RQ3: To what extent do LLM-generated responses exhibit appropriate contextual and cultural grounding when addressing crisis scenarios in Ukrainian compared to English baselines?}

Generally, LLMs have been shown to face challenges in Contextual \& Cultural Grounding \cite{rystrom2025multilingual}, with this metric being much more specific and tailored to human perception, rather than a capability which can be directly inherited from digitized corpora alone. Therefore, Gemini and LLaMA experience difficulties in adapting effectively to users' emotional states, as is evident from the Human Evaluation Baseline. Empirical evidence suggests that English-language responses target a much broader audience \cite{grattafiori2024llama3herdmodels,
deepseekai2025deepseekv3technicalreport}, therefore making it easier to address queries using general idioms of distress. Nevertheless, the Ukrainian language represents a much narrower community than the English baseline does, making it harder to address the need for Emotional \& Cultural Grounding.

As discussed in the previous sections, the decline in one metric often coincides with lower performance in the others due to the close relationship between these dimensions. When LLaMA may experience difficulties in producing more natural phrasing because of language drifting and byte-fallback phenomena when answering in Ukrainian, it directly affects its ability to provide grounding to an individual, substantially undermining the perception of a natural conversation. Similarly, if Gemini's phrasing contains over-alignment bias or becomes lexically saturated, its Empathetic Accuracy may decrease as well. While some grounding phrases are appropriate in an English-speaking environment, their direct translation into Ukrainian causes awkward, unnatural, and robotic phrasing that may not be fully noticeable to a machine.

However, while LLMs' multilingual abilities might be far from perfection, the distinction between multilingual competence and Contextual \& Cultural Grounding becomes particularly evident from the perspective of native Ukrainian speakers when compared to English. Therefore, while models can possess multilingual competence, Cultural Grounding requires a deep understanding of norms, values, emotional expectations, communicative conventions, and implicit references \cite{rystrom2025multilingual}. Consequently, both prior research and the findings obtained through the SPLIT benchmark suggest the following conclusion:

\begin{quote}
    \centering
    \textit{Producing Ukrainian text is not equivalent to producing Ukrainian emotional support.}
\end{quote}

In the context of the current conflict, it is particularly important to understand the cultural and contextual nuances associated with the Ukrainian language, as linguistic fluency alone is not the most challenging component of effective communication. Nevertheless, prior research consistently indicates that English training corpora remains  dominant within the LLM paradigm \cite{grattafiori2024llama3herdmodels,
deepseekai2025deepseekv3technicalreport}. This  may therefore lead to excessive formality and general reassurance, the exact tendency observed in LLaMA and Gemini models, consequently resulting in lower scores.

Interestingly, DeepSeek's performance differs across all three parameters of our SPLIT benchmark. As has already been explained above, DeepSeek's internal architecture and both its pre-training and post-training objectives may have contributed to its strong performance in both English and Ukrainian samples of users' queries. The relatively high scores assigned to the model appear to reflect its ability to maintain emotional continuity throughout responses. While not exhibiting sufficiently human-like performance to receive perfect scores, its ability to maintain stable performance when transitioning to a less represented language constitutes a notable finding in itself.

\subsection{RQ4: To what extent does automated LLM-based evaluation agree with human assessment of empathetic conversational responses?}

Stating that there is no agreement between the two evaluation baselines, human and AI, would be statistically incorrect, as demonstrated by the Pearson correlation coefficient calculated and examined in the Results and Analysis section. Accordingly, the Empathetic Accuracy metric demonstrates weak, but highly significant alignment, suggesting that AI systems can approximate certain aspects of human judgments of empathy \cite{paech2024eqbench, malik-etal-2025-llms}.

This is additionally evident from the Grand AI Averages Baseline, with the scores for English responses showcasing a similar tendency. However, while the automated jury is capable of capturing general patterns of reassurance suited for dominant English-speaking users, the scores never dropped significantly in Ukrainian response evaluations, indicating limitations in the jury's performance in this context. Specifically, Empathetic Accuracy is partially overscored, with positive $MAE$ and $ME$ values, suggesting the presence of Leniency Bias. This reinforces an interesting observation, where AI tends to assign higher scores to potentially empathetic responses containing basic, though emotionally appropriate, vocabulary \cite{paech2024eqbench, malik-etal-2025-llms}.

Despite the challenges LLMs face when addressing emotionally draining situations, the score for DeepSeek remains relatively stable for both English and Ukrainian. It may indicate that AI has the ability to detect more relevant, natural idioms of distress. However, considering the similar scores assigned to the other two assessed models, which do not change significantly with the change of language, this posits the idea that the LLM-as-a-jury paradigm may rely on surface-level linguistic indicators rather than deeper emotional representations of empathy.

Linguistic Naturalness demonstrates a similar pattern of approximately the same correlation, but with slightly weaker significance, though still sufficient enough to showcase a persistent level of alignment with the Human Evaluation Baseline. However, the $MAE$ and $ME$ values presented in the section above indicate that the magnitude of bias is considerably larger, suggesting that AI and human evaluators attend to different characteristics of the generated responses. One of the most reasonable explanations is that the AI jury rewards structure and grammar more than authenticity \cite{xing2024evaluating, 
han2025mubenchassessmentmultilingualcapabilities}, thus reflecting patterns inherited from the training corpora of the evaluating models.

With the Contextual \& Cultural Grounding dimension being more complicated to assess, even for human beings, the LLM-as-a-jury framework exhibits no significant alignment, thus demonstrating a negative correlation and low significance, showing no meaningful agreement between human and the LLM-as-a-jury paradigm. Although both humans and LLMs recognize certain aspects of cultural grounding, according to the baseline averages, LLMs appear to demonstrate limited sensitivity to cultural nuances and implicit meanings \cite{rystrom2025multilingual}. It further leads to a broader conclusion: Large Language Models demonstrate strong statistical performance for simpler, less complex dimensions that rely less heavily on culturally embedded knowledge, which may not be fully captured by training corpora alone. This deduction is also justified by slightly negative $ME$ and positive $MAE$ values, supporting the interpretation that AI and human juries face difficulties in applying the same evaluation criteria to this dimension, with some words carrying implicit, culturally appropriate meanings.

Overall, the results of our LLM-as-a-jury framework can be summarized as follows:

\begin{quote}
    \textit{Automated evaluation may reliably capture certain superficial dimensions of empathetic communication, while diverging from human perception in aspects requiring contextual and cultural immersion.}
\end{quote}

\newpage
\section{Limitations}

Our study specializes in a niche subject, consequently aiming to explore and contribute to the realm of Large Language Models and cross-lingual inconsistencies encountered when addressing crisis-related queries by introducing our own SPLIT benchmark. Nevertheless, several limitations should be acknowledged:

\subsection{Human Annotation Subjectivity}

First and foremost, the Human Evaluation Baseline established in our study relies specifically on a single evaluator, meaning that the outcome may remain relatively subjective. Accordingly, some degree of disagreement would likely emerge if the same responses were evaluated by another annotator.

To address this specific issue, future work in this sphere should adopt an inter-rater agreement metric to average the outcomes and produce more culturally reliable results. Nevertheless, the current study aims to provide an unbiased assessment, informed by  relevant peer research and a range of other considerations regarding the deployed models established preliminarily.

\subsection{Language Scope}

While our findings resemble those found in a range of peer research, they do not observe empathy divergence across all low-to-mid resource languages. While some broad statements are employed, their purpose is to state the omnipresence of the current cross-lingual issue, rather than claim or appropriate other studies' results. Further research should implement a number of other less represented languages to explore the lack of cultural understanding in LLMs more rigorously and grasp the full scope of such a multifaceted issue.

\subsection{Benchmark Scope}

The SPLIT benchmark specializes in testing queries in five specific topics, such as Stress, Panic, Loneliness, Internal Displacement, and Tension. Therefore, a range of other potential bottlenecks remains to be included, with ours covering only several of them, which are directly related to crisis-affected scenarios in a Ukrainian context. We do not state that LLMs' lack of cultural understanding directly transfers to other areas, such as education, healthcare, business communication, or general dialogue. These are separate subjects, which are intentionally not included in our benchmark.

\subsection{Statistical Scope}

The SPLIT benchmark encompasses 500 crisis-related queries, with 100 in each of the evaluated topics. In total, 3000 responses are generated by three LLMs, 10\% of which are manually evaluated. While this sample is sufficient for exploratory analysis, it may not capture all forms of scenarios. Therefore, future work could incorporate additional reliability metrics or simply extract a larger sample for human validation.

\subsection{Model Selection}

Our research employed more than one, specifically three LLMs to address the queries generated for our SPLIT benchmark. Additionally, three completely different models in terms of architecture were deployed to mitigate bias levels in the assessments and make our research more valid and multifaceted. Nevertheless, to ensure higher credibility, future work should aim to adopt a more comprehensive approach, using a wider range of models for both answer generation and their assessment.

Furthermore, proprietary models evolve rapidly; thus, future versions might exhibit different behavior, consequently altering the models' answers and assessments.

\section{Conclusion}
We introduce the SPLIT benchmark, aimed at evaluating Cross-Lingual Empathy Divergence and Conversational Discrepancies in English and Ukrainian LLM Responses. Our 500-prompt dataset consists of 500 Ukrainian crisis-related scenarios across five categories: Stress, Panic, Loneliness, Internal Displacement, and Tension. Our study combines prior peer research, the assessment of three LLMs, and the establishment of the LLM-as-a-jury framework. Our work encompasses both human and automated assessment, additionally aiming to validate LLMs' capabilities in detecting cultural and linguistic nuances. Thus, we evaluate three dimensions: Empathetic Accuracy, Linguistic Naturalness, and Contextual \& Cultural Grounding. 

Our empirical findings reveal that several  models, such as Gemini-2.5-Flash and LLaMA-3.3-70B-Instruct experience degradation in the Empathetic Accuracy and Linguistic Naturalness metrics, while DeepSeek-V3 demonstrates stability when addressing crisis-related and empathy-demanding queries within our SPLIT benchmark. However, a positive correlation suggests that a certain degree of agreement exists between human and AI assessments. Having conducted a thorough analysis, a conclusion has been made: \textit{fluency does not necessarily imply naturalness}. In this way, human evaluators tend to penalize translated, rigid, or overly formal expressions much more strongly than AI judges. 

Contextual \& Cultural Grounding emerges as the most challenging dimension, showing negative, insignificant alignment between human and AI evaluators, suggesting that the LLM-as-a-jury paradigm struggles to apply evaluation criteria that align with human judgments for this dimension. Therefore, while showcasing decent grammatical structures, the models are not necessarily able to follow culturally tailored crisis-related queries. This has led to the core final idea derived from evaluating the results of our SPLIT benchmark: \textit{producing Ukrainian text is not equivalent to producing Ukrainian emotional support.} Therefore, automated evaluation may capture some dimensions reliably, while diverging from human perception in dimensions requiring cultural and contextual immersion.

Our findings aim to assist in the further development of LLMs and their implementation in emotional-support-demanding situations. We reinforce the idea that LLMs deployed in crisis contexts require deeper cultural adaptation. Existing evaluation benchmarks may overlook the Cultural Grounding dimension, consequently demonstrating the need for multicultural, not merely multilingual, benchmarks. Specifically, LLM developers should take into account the cultural nuances of specific mid-to-low-resource languages when cultivating and adopting post-training objectives.

Ultimately, as LLMs are increasingly integrated into crisis response systems, ensuring they possess genuine cultural immersion, rather than simply linguistic fluency, is no longer just an optimization goal, but a fundamental requirement for safe and responsible deployment.

\section{Acknowledgments}
The author wants to thank Professor Russell Reid for valuable comments and suggestions on an earlier version of this paper. Any remaining errors or inaccuracies are solely the author's responsibility.

\bibliography{references}

\end{document}